\pdfoutput=1

\documentclass[11pt]{article}

\usepackage{ACL2023}

\usepackage{times}
\usepackage{latexsym}

\usepackage[T1]{fontenc}

\usepackage[utf8]{inputenc}

\usepackage{microtype}
\usepackage{arydshln}
\usepackage{multirow}
\usepackage{tikz}
\usepackage{pgfplots}
\usepackage{amsmath}
\usepgfplotslibrary{external}
\usepackage{algorithm}
\usepackage{algpseudocode}

\usepackage{inconsolata}

\title{Self-generated Replay Memories for Continual Neural Machine Translation}

\author{Michele Resta$^{\spadesuit}$  \quad Davide Bacciu$^{\spadesuit}$ \\
         $^{\spadesuit}$ University of Pisa, Largo B. Pontecorvo, 3, 56127, Pisa, Italy \\  \texttt{michele.resta@phd.unipi.it} \quad \texttt{davide.bacciu@unipi.it}}

\begin{document}
\maketitle
\begin{abstract}
Modern Neural Machine Translation systems exhibit strong performance in several different languages and are constantly improving. Their ability to learn continuously is, however, still severely limited by the catastrophic forgetting issue.
In this work, we leverage a key property of encoder-decoder Transformers, i.e. their generative ability, to propose a novel approach to continually learning Neural Machine Translation systems. We show how this can effectively learn on a stream of experiences comprising different languages, by leveraging a replay memory populated by using the model itself as a generator of parallel sentences. We empirically demonstrate that our approach can counteract catastrophic forgetting without requiring explicit memorization of training data. Code will be publicly available upon publication\footnote{\url{https://github.com/m-resta/sg-rep}}.
\end{abstract}

\section{Introduction}

Neural Machine Translation (NMT) systems have achieved remarkable performance on numerous language pairs, particularly those with abundant resources. The substantial growth in model parameters and the availability of large crawled corpora have greatly contributed to the adoption of advanced techniques like back-translation \cite{edunovUnderstandingBackTranslationScale2018, DBLP:conf/acl/SennrichHB16} and denoising pre-training \cite{mbart, mass, xue-mt5}, further increasing the translation quality of these models. Consequently, even low-resource languages have benefited from the increased multilingual capabilities of these models \cite{massive-mnmt, m2m-100}.

Despite their impressive quality, modern NMT systems exhibit limited Continual Learning (CL) capabilities and are susceptible to catastrophic forgetting (CF, \citealp{MCCLOSKEY1989109, french2006}). While the CL community has extensively investigated this phenomenon over the years, it has received relatively less attention from the Natural Language Processing (NLP) community. Previous works in this area have primarily focused on Domain Adaptation, and only recently have turned to different continual learning settings such as incremental language learning \cite{cill-bench}.

In this paper, we present \textbf{SG-Rep}: a novel approach to continually train an NMT system on a stream of experiences comprising several language pairs while mitigating the detrimental effects of catastrophic forgetting. Our method leverages a replay memory populated by synthetic parallel sentences generated by the model itself. We evaluate the effectiveness of our approach across different translation directions and demonstrate its ability to alleviate CF without the need for explicit memorization of the training data.
This aspect is crucial where is not possible to store real training samples for privacy reasons or data retention policies.

\begin{figure}[]
    \centering
    \includegraphics[scale=0.3]{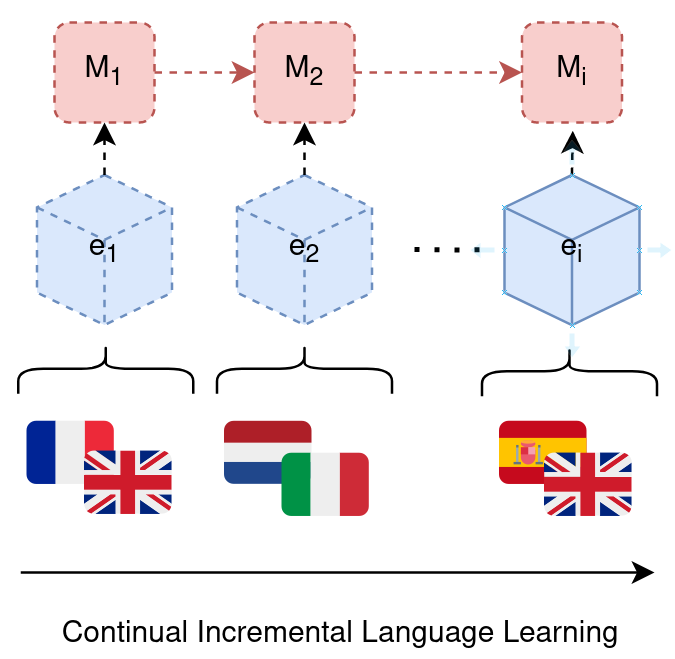}
    \caption{A scheme of the CILL setting. A model is trained incrementally on a stream of experiences comprising training data for various language pairs.}
    \label{fig:cill}
\end{figure}

\section{Related Works}

Continual Learning or Lifelong Learning (LL) research focuses on developing computational systems that can gradually acquire, refine, and transfer knowledge over extended periods, imitating biological systems. 
The primary challenge arises from the inherent plasticity of neural networks, which leads to catastrophic forgetting. This refers to a situation where performance on previously learned tasks deteriorates as the network parameters are updated, resulting in the loss of acquired knowledge.

In addition to mitigating CF, lifelong learning methods should also promote knowledge transfer across tasks in both forward and backward directions.
Research work in the field can be broadly categorized into architecture, regularization, or data-based approaches.

\textbf{Architecture-based Methods}

The central idea of this category of approaches is to allow the network architecture to change either by adding new parameters for specific tasks or by maintaining a fixed network size while allocating a different capacity to each task.
Progressive networks \cite{prog-net} maintain a pool of pre-trained models throughout training and learn lateral connections to leverage useful features for the new task.
\citealp{DBLP:journals/neco/SodhaniCB20} combine Net2Net \cite{DBLP:journals/corr/ChenGS15} (a network growing approach) together with a gradient episodic memory to enable RNNs to dynamically expand if they fail to learn the current task.

PathNet \cite{Pathnet} presents an evolutionary-based algorithm that identifies which parts of the network to reuse for new tasks. By preserving parameters along a learned path from task A and evolving new paths for task B, accelerated learning is achieved compared to starting from scratch or fine-tuning.

Examples of architectural methods in NLP include instantiating a new decoder module for learning new translation directions in NMT \cite{DBLP:conf/acl/EscolanoCF19}. Another approach is the use of small task-specific modules called "adapters" \cite{DBLP:conf/nips/RebuffiBV17, DBLP:conf/icml/HoulsbyGJMLGAG19, adapterfusion} to avoid finetuning of large pre-trained models.

\citealp{CL-MNMT-berard} focuses on learning a set of language-specific embeddings to be used at inference time to boost translation quality.
Other works \cite{DBLP:conf/aaai/LiangZWQ021, DBLP:conf/naacl/GuFX21} have explored pruning as a parameter partition strategy.

In contrast, \citealp{cao-cl}. propose a vocabulary-based approach to address CF, utilizing a vocabulary adaptation scheme that leverages the token overlap in the vocabulary of M-NMT systems that support multiple languages. When a new language is introduced, a new vocabulary is created, encompassing all the data. Embeddings for tokens in the intersection of the old and new vocabularies are reused, and training continues.

We identify two main limitations: first, training a new vocabulary requires storing all data up to the current experience, second, the starting vocabulary must support a large number of languages (24 in the paper) to achieve significant token overlap with the newly trained vocabulary.

\textbf{Regularization-based Methods}

This family of methodologies is based on theoretical neuroscience models that suggest synapses with varying levels of plasticity can protect acquired knowledge. In computational terms, this is implemented by adding a regularization term to the loss function of a neural network, leading to constrained weight updates.
Elastic Weight Consolidation (EWC) \cite{EWC} prevents catastrophic forgetting by slowing down learning on important weights for old tasks, using the Fisher information Matrix 
to estimate their importance. 

Memory Aware Synapses (MAS) \cite{MAS} computes parameter importance in an unsupervised manner so that they approximate the sensitivity of the learned function to a parameter change.
For domain adaptation, \citealp{overc-thompson} employs EWC as a regularizer, while \citealp{DBLP:conf/naacl/LiCCHLXGL22} propose an approach based on estimating the domain shift.
\citealp{DBLP:conf/aclnmt/KhayrallahTDK18} use a Knowledge Distillation (KD) inspired regularization approach.
\citealp{shao-cl} presents Online Knowledge Distillation (COKD) with complementary training and KD, distilling knowledge from n teachers to the student. \citealp{cao-cl} proposes Dynamic KD, optimizing a weighted sum of translation and distillation loss, tested with fixed language pairs in both domain and time incremental fashions. Both COKD and Dynamic KD are resource-intensive, relying on external models for distillation, in contrast to our approach. Additionally, \citealp{cao-cl} tested on fewer experiences.

\textbf{Data-based Methods}

Data-based methods retain a small number of training samples from previous tasks to limit weight updates based on the data distribution of past experiences. These samples can be either real or pseudo samples.

In the case of real samples, the work of \citealp{DBLP:conf/acl/ChuDK17} demonstrates the use of a mixture of old and new data during domain adaptation. Additionally, the incorporation of replay data and retrieval mechanisms has been shown to improve performance in neural machine translation \cite{DBLP:conf/naacl/BapnaF19, DBLP:conf/acl/XuCS20}.
GEM \cite{GEM} and A-GEM \cite{A-GEM} utilize real samples from previous tasks to constrain gradient updates within a favorable region for the current task, leading to better performance.

Other approaches closer to our proposed methods, leverage generative models to create pseudo samples: LAMOL   \cite{LAMOL} and   \cite{DBLP:conf/acl/Zhang0Y22, DBLP:conf/iclr/QinJ22} are among the methods that adopt this strategy.

\section{The Self-Generated Replay Method}
The goal of our method,which we abbreviate as \textbf{SG-Rep}, is to incrementally learn a single model that is able to translate into many directions. The training data is not available as a whole but is presented to the model in incremental steps, as a stream of experiences that comprises one or more language directions. The key challenge is preserving translation performance on past experience while adapting effectively to new data. To this end, we use the encoder-decoder model learned in a given experience as a generator of synthetic training samples.
The obtained pseudo-samples will populate a fixed-size replay memory that will be used in future tasks to mitigate catastrophic forgetting.

\subsection{Continual Incremental Language Learning}
\label{sec:cill}
Similarly to other works from the CL community, we design a setting where the learning process is divided into $E$ learning experiences.  In each $e_i \in E$ the model is exposed to a pair of languages $l_1^i, l_2^i$ and has access to a training set $T_i$ and a validation set $V_i$ comprising a single or both translation directions.
In the Continual Incremental Language Learning (CILL) scenario (Figure \ref{fig:cill}) the model is trained across all experiences: we want to model  $P ( \textbf{y} | \textbf{x}) $ for all the languages of interest, with $\textbf{x}$ and $\textbf{y}$ being the source and target sentence respectively. The model architecture and the number of parameters are kept fixed, together with the sub-word vocabulary which is built in advance. 
We design a set of experiments to quantify the amount of catastrophic forgetting occurring during the subsequent experiences and the effectiveness of our proposed replay strategy.
We employ a fixed-size memory (the replay buffer) that is filled at the end of each $e_i$ experience. 
We experimented with different buffer sizes and performed also experiments without memory, to assess the amount of catastrophic forgetting. In all the experiments the replay buffer size is fixed and it is not allowed to grow.

\subsection{Self-generated Replay Memories}

SG-Rep generates pseudo samples following three main steps: 1) generation of samples, 2) filtering, and 3) samples translation. 

\textbf{Encoder input.} Consider $l_s \rightarrow l_t$ as translation direction, and a model $M$ capable of translating in both directions: $l_s \leftrightarrow l_t$.

In the initial stage, our method generates a sentence in language $l_t$ by using as encoder input a short text $t_e$ that contains a special token indicating the translation direction \texttt{<2lang>}. Here, "lang" represents a 2-letter language code that identifies the target language $l_t$.
We experimented with concatenating $t_e$ with random words to improve input diversity but we observed a detrimental effect on the overall performance.

\textbf{Generation and Filtering.}
We generate $n$ pseudo sentences in language $l_t$ iteratively by top-k sampling with $k$ equal to vocabulary size. 
At each step, we process the sentences by  1) eliminating duplicates and 2) filtering out low-quality ones.

The filtering criterion considers morphological correctness. We use the PyEnchant\footnote{https://github.com/pyenchant/pyenchant} spellchecker to identify the number of misspelled words $\hat{w}$ in each sentence and filter out those with $\hat{w} \geq 2$.

\textbf{Translation.} The pseudo sentences are then translated into language $l_s$ by using the same model, obtaining a set of self-generated samples $R = \{(x_i,y_i)\}_{i=1}^n$ with $x$, and $y$ in language $l_s$ and $l_t$ respectively.

The last step populates the replay memory $R^*$ by reservoir sampling from $R$. The total amount of samples in the replay buffer is constant during all the experiences, while their proportions vary for the effect of the sampling. The training samples for the next experience $e_i$ will then be $T_i \cup R^*$. We repeat the process for each translation direction present in the current experience.  The pseudo-code of SG-Rep is reported in Algorithm \ref{alg:self-gen}.

\begin{algorithm}[tb]
        \begin{algorithmic}[1]
            \Statex \textbf{Input}:
            \Statex \hspace*{\algorithmicindent}$M$: NMT model
            \Statex \hspace*{\algorithmicindent}$n$: \# samples to generate
            \Statex \hspace*{\algorithmicindent}$l_s, l_t$: source and target language
            \Statex \textbf{Output}:  $R$: list of source-target pairs
            \Procedure{generate\_replay\_data}{}
                
                \State $R\gets$  [ ] 
                \State $t_e \gets $\Call{get\_encoder\_input}{$l_t$}
                \State $src\gets$  [ ] \Comment{Pseudo-source sentences}
                \State $tgt\gets$  [ ] \Comment{Translated sentences}
                \While {$length(src)$ < $n$}
                    \State $out\gets$  \Call{generate}{$t_e$, $M$}
                    \State $out\gets$  \Call{filter}{$out$}
                    \State $tgt\gets$  $tgt$ + $out$
                    \State $tgt\gets$  \Call{deduplicate}{$tgt$}
                \EndWhile

                \State $src\gets$   \Call{translate}{$tgt$, $l_s$, $M$} \Comment{Translate $src$ into source language $l_s$}
                \State $R\gets$  $[src, tgt]$
                \State \Return $R$
            \EndProcedure
        \end{algorithmic}
        \caption{SG-Rep pseudocode}
        \label{alg:self-gen}
\end{algorithm}  

\section{Experimental Setting}

\subsection{Multilingual Translation Model}
\label{sec:model_arch}
A single Transformer model \cite{vaswani-transformers} was chosen as the architecture of the multilingual NMT system. 
We started from the \textit{T5 small v1.1} described by Google \cite{t5}  and available via Huggingface's Transformer library. We reduced the number of encoder and decoder blocks to 6, and the number of attention heads to 8 in both the encoder and decoder. The total amount of parameters with this configuration is $54.15 \cdot 10^6$. Before training all weights were reinitialized.

 We prepend language tokens to the source sentences to denote the desired translation direction as in \cite{massive-mnmt}. 
The SentencePiece \cite{sentencepiece} tokenizer is trained with a minimum merge frequency of 5 and a vocabulary size of 32k tokens by using HuggingFace \cite{huggingface}.
We trained in advance a total of 3 tokenizers: two for the two sets of languages from IWSLT17 and one for those from UNPC.

\subsection{Datasets and experiences}
We run experiments with both small and large-scale datasets. In the small scale setting  we employ a subset of the IWSLT17 \cite{iwslt17} dataset while for the large scale one we resort to the United Nation Parallel Corpus (UNPC) \cite{unpc-corp}. 
For all the experiments the chosen language pairs are organized into four bidirectional experiences. In each experience, the model is exposed to a language pair and learns to translate from the source language to the target language and vice versa. This choice is motivated by two main reasons. Firstly, the inherent bidirectionality of the translation task. Secondly, in principle, in future steps, we may not have access to the training samples of a particular experience. Therefore, by alternating the roles of source and target languages in the training samples, we aim to ensure the model receives as much relevant information as possible.

\textbf{IWSLT17.} To create the first stream we focus on the following translation directions: French $\leftrightarrow$ English, Dutch $\leftrightarrow$ Italian, English $\leftrightarrow$ Romanian, and Italian $\leftrightarrow$ Romanian.
The second stream contains also non-European languages: Arabic $\leftrightarrow$ English, English $\leftrightarrow$ French, Korean $\leftrightarrow$ English, and Italian $\leftrightarrow$ Dutch.

\textbf{UNPC.} In this dataset we focus on a mix of European and non-European languages: Arabic $\leftrightarrow$ English, Spanish $\leftrightarrow$  Russian, and English $\leftrightarrow$ French.

Table \ref{tab:experiences} summarizes the experiences and their corresponding data sizes for all streams.

 \begin{table}[]
 \centering
 \resizebox{0.35\textwidth}{!}{\begin{tabular}{ccrrr}
 \hline
 \multicolumn{5}{c}{\textbf{IWSLT17}}\\
 \textbf{Exp.} & \textbf{Direction} & \# \textbf{Train} & \# \textbf{Dev} & \# \textbf{Test} \\
 \hline
 1   & Fr $\leftrightarrow$ En    & 465,650   & 1,780  & 17,194    \\
 2   & It $\leftrightarrow$ Nl    & 466,830   & 2,002  &  3,338    \\
 3   & En $\leftrightarrow$ Ro    & 441,076   &  1,828 &  3,356    \\
 4   & It $\leftrightarrow$ Ro    & 435,102   & 1,828   &  3,268  \\
 \hline
 \hline
 1   & Ar $\leftrightarrow$ En    & 463,426   & 1,776  & 17,166    \\
 2   & En $\leftrightarrow$ Fr    & 465,650   &  1,780 &  17,194    \\
 3   & Ko $\leftrightarrow$ En    & 460,480   &  1,758 &   17,028   \\
 4   & It $\leftrightarrow$ Nl    & 466,830   &  2,002 &  3,338  \\
 \multicolumn{5}{c}{\textbf{UNPC}}\\
 \hline
 1   & Ar $\leftrightarrow$ En    & 20,040,478 & 4,000   & 4,000    \\
 2   & Es $\leftrightarrow$ Ru    & 22,290,106 & 4,000   & 4,000    \\
 3   & En $\leftrightarrow$ Fr    & 30,336,652 & 4,000   & 4,000    \\
 3   & En $\leftrightarrow$ Es    & 25,223,004 & 4,000   & 4,000    \\

 \hline
 \end{tabular}}
 \caption{Summary of the different streams of experiences. IWSLT17: the first one (top) is composed of European-only languages while the second one (bottom) contains also non-European ones. UNPC: The four experiences contain European and non-European languages. We report the total size of train, validation, and test datasets. In a single experience, each direction has the same amount of samples for each split.}
 \label{tab:experiences}
 \end{table}

\subsection{Systems}
We evaluate different systems against our proposed self-replay approach (\textbf{SG-Rep}). Each system follows the architecture described in Section \ref{sec:model_arch} and has been implemented using the Transformers library \cite{huggingface} except where explicitly indicated.
\begin{itemize}
    \item \textbf{SG-Rep (ours).} We fix top-$k$ sampling  temperature at $T=0.93$. The translation phase is conducted with a beam search approach with a beam size of 12. 
    The replay buffer size is always computed based on the total training data for the first IWSLT17 stream (the one with European-only languages). This allows us to have a relatively small memory also for experiments with large corpora and to compare results. We varied both the buffer sizes and the amount of generated samples.  The latter is indicated with a superscript. SG-REP$_{0.05}^{100}$ denotes a buffer size of 5\% of the data randomly filled using 100K generated translation pairs.
    Once set, the buffer size is not allowed to increase and stays constant throughout the experiences.
    We explicitly report the different sizes of the memory in Appendix \ref{appendix:mem-size}.
    \item \textbf{Incremental Training.} The model is directly trained on each experience, one after another.
    \item \textbf{Multitask (Joint training).} The classical upper bound for CL: the model is trained on data from all experiences simultaneously. 
    \item \textbf{Replay.} The model has a fixed memory buffer. At the end of each experience, the memory is randomly populated with examples from the training set of the current experience using reservoir sampling. We tested different memory sizes, expressed as a percentage as we did for SG-Rep. We indicate the buffer percentage with a subscript: Replay$_{0.05}$ denotes a 5\% replay memory buffer.
    \item \textbf{EWC.} The model uses Elastic Weight Consolidation as a regularization strategy. EWC computes the importance of the parameters using the Fisher information matrix and penalizes changes to important ones so that they stay close to the original ones. The regularized loss function is:
    \begin{equation}
    \mathcal{L}(\theta) = \mathcal{L}_{CE}(\theta) + \frac{\lambda}{2}\sum_i F_i(\theta_i -\theta_{0,i}^*)^2
    \end{equation}
    with $\lambda$ controlling the relevance of the old task compared to the new one. We follow the implementation from the Avalanche library \cite{avalanche}. 
    \item \textbf{A-GEM.} This GEM followup \cite{GEM} uses a small episodic memory to store a subset of the examples from each experience. When training on subsequent tasks, a random batch is sampled from the memory, and the losses on the episodic memories are treated as inequality constraints. A-GEM tries to ensure that the average episodic memory loss over the previous tasks does not increase, and improves on GEM computational requirements. We set the batch size equal to 150 when sampling from the memory. The implementation follows the one from the Avalanche library \cite{avalanche}.
    \item \textbf{LAMOL.} A language model is treated as both the learner and the generator. The model is trained by casting different tasks to question answering (QA) in a SQuaD \cite{squad} format. During training, each example is formatted into both the QA format and the language modeling (LM) format.
    The total loss optimized by the model is a weighted sum of the QA loss, and the classical LM loss:
    \begin{equation}
        \mathcal{L}(\theta) = \mathcal{L}_{QA}(\theta) + \lambda \mathcal{L}_{LM}(\theta).
    \end{equation}
    We use the authors' original implementation after pre-processing the IWSLT17 dataset in the required format. 
    We ran experiments with both GPT and GPT-2 \cite{gpt-2} models with a total of 116M and 124M trainable parameters, respectively. 
\end{itemize}

\subsection{Training Details} 
We train all configurations for a maximum of 50 epochs with early stopping (patience = 10) using the Adam optimizer ($\beta_1=0.9$, $\beta_2 = 0.99$). We adjust the learning rate with a cosine scheduler ( $warmup\_steps = 16k$). We evaluate the models every $5k$ steps, starting with an initial learning rate of $5 \cdot 10^{-4}$ and apply a dropout rate of 0.1. For decoding, we use beam search (size 12) with a maximum length of 128 tokens. We train all systems on a single A100 GPU, with a batch size of 150 in FP16 precision.For LAMOL, we train both the GPT and GPT-2 models using the author's code and the hyperparameters reported as the best in their paper. We train for 9 epochs on 8 H100 GPUs with a batch size of 310.

\section{Experimental Results}

The translation output of the models is scored using the 4-gram
case-sensitive BLEU \cite{BLEU} with the SacreBLEU tool \cite{sacrebleu} using the default  tokenization scheme\footnote{BLEU+case.mixed+numrefs.1+smooth.exp+tok.13a +version.2.3.1} based on mosestokenizer\footnote{https://github.com/moses-smt/mosesdecoder/blob/master/scripts/tokenizer/tokenizer.perl}. 
COMET scores are reported in Appendix \ref{appendix:comet-scores}.

Each model is evaluated at the end of training on the last experience on all translation directions. 
In addition to the average BLEU, we also report a  metric for comparative analysis to compensate for the diversity of the different test sets for each language. It quantifies the performance delta between the chosen system and the upper bound (i.e. joint training) for each language pair.
Specifically, we define the language pair delta as $$\Delta Lp^* = \sum_{i=0}^n U_i - S_i $$ where $U_i$ and $S_i$ denote the BLEU scores for language pair $i$ under the upper bound and the system under consideration, respectively and $n$ is the number of language pairs. We indicate with '*' the system representing the upper bound.
Additionally, we conducted an analysis on data leakage and generated pseudo samples that we report in Section \ref{sec:data-leak} and Appendix \ref{appendix:true-vs-false}, respectively.

\subsection{IWSLT17 European Languages only}

\begin{table*}[]
\centering
\resizebox{0.85\textwidth}{!}{\begin{tabular}{l|cc|cc|cc|cc|cc}
\hline

\textbf{Systems } & \textbf{Fr$\rightarrow$En }     & \textbf{En$\rightarrow$Fr}      & \textbf{Nl$\rightarrow$It }      & \textbf{It$\rightarrow$Nl}       & \textbf{En$\rightarrow$Ro}       & \textbf{Ro$\rightarrow$En }     & \textbf{It$\rightarrow$Ro }      & \textbf{Ro$\rightarrow$It}      & \textbf{Avg.}  & $\mathbf{\Delta Lp^*} \downarrow $   \\
\hline
\hline
Incremental train    &  0.44  & 0.82 &       1.58 &   1.05 &  17.92 &   0.97 &  24.14 &  23.11 &  8.75 & 20.48 \\[4pt]
EWC   & 1.32 &   1.12 &  12.41 &   1.08 &  22.02 &   1.12 &  20.60 &  22.18 &   10.23 & 19.00\\[4pt]
LAMOL$_{\text{GPT-2}}$  &  11.69 &   5.54 &   3.14 &   1.93 &   9.66 &  12.83 &   6.52 &   6.24 &  7.19 & 22.04\\[4pt]
LAMOL$_{\text{GPT}}$      &  15.27 &  12.55 &   6.20 &   4.89 &  18.39 &  18.97 &  12.37 &  10.81 &  12.43 &16.80\\[4pt]
A-GEM$_{0.2}^{}$                   &  \textbf{33.81} &  \textbf{31.06} &   1.48 &   0.95 &  23.69 &   4.63 &  21.36 &  23.51 &  17.56 & 11.67 \\[4pt]
SG-Rep$_{0.1}^{180}$ &  24.70 &  20.84 &  16.74 &  13.18 &  26.48 &  23.00 &  24.44 &  23.51 &  21.61 & 7.62\\[4pt]
SG-Rep$_{0.2}^{100}$  &  28.65 &  25.35 &  17.08 &  15.13 &  24.68 &  \textbf{26.04 }&  21.41 &  23.04 &  22.67 & 6.56 \\[4pt]
SG-Rep$_{0.2}^{180}$ &  27.51 &  24.92 &  18.01 &  14.94 &  26.81 &  25.71 &  \textbf{24.57} &  \textbf{23.38} &  23.23  & 6.00 \\[4pt]
SG-Rep$_{0.2}^{250}$        &  29.00 &  26.65 &  \textbf{18.11} &  \textbf{15.06} &  \textbf{28.00} &  24.91 &  24.36 &  23.17 &  \textbf{23.66} & \textbf{5.57} \\[2pt]
\hdashline
Replay$_{0.1}^{}$            &  36.06 &  33.92 &  19.94 &  19.45 &  29.79 &  31.59 &  24.27 &  23.34 &  27.30 & 1.94 \\[4pt]
Joint Training*          &40.60 &  39.47 &  21.98 &  22.03 &  28.17 &  35.10 &  21.95 &  23.66 &  29.12 & --\\
\hline
\end{tabular}}
\caption{Scores of the different methods evaluated at the end of the last experience. The Avg column is the average BLEU score across all translation directions. Joint training and Replay are shown as upper bounds at the bottom.}
\label{tab:main_results}
\end{table*}

Our main results are summarized in Table \ref{tab:main_results}.
As can be seen, incremental training results in extreme catastrophic forgetting causing the model to completely lose its translation capabilities with the exception of languages present in the last experience. 
EWC brings only a slight performance improvement over fine-tuning in this training scenario.

All data-based methods perform better than EWC.
Surprisingly, in this experimental setting, A-GEM surpasses LAMOL. We hypothesize that this result is due to the difficulty of the task as LAMOL was originally designed to deal with training data cast as QA. Compared to the original setting, casting a translation task as QA will yield longer answers, and consequently, increased generation difficulty.
Thus the external memory used by A-GEM proves advantageous.

SG-Rep has the best overall performance among all tested methods and is the one getting closer to both joint training and replay using real samples.

\begin{figure}[h]
    \centering
    \includegraphics[width=0.9\columnwidth]{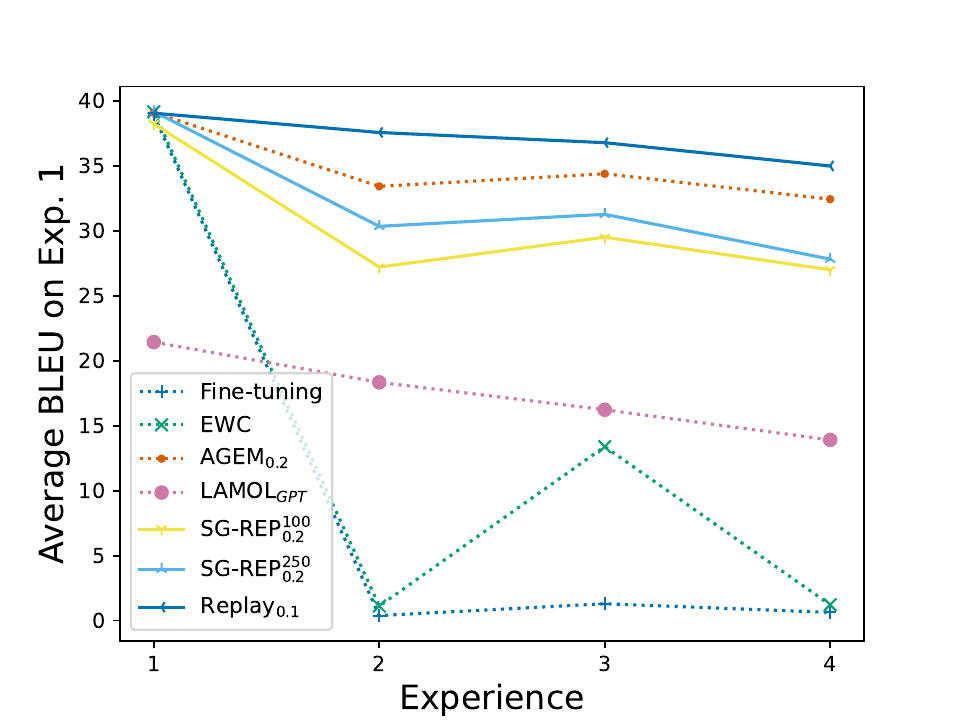}
    \caption{Forgetting curve of the different approaches. Average BLEU score on the first task evaluated at the end of the training process of each experience.  }
    \label{fig:forgetting}
\end{figure}

Figure \ref{fig:forgetting} shows the forgetting curve of the various baselines: we compute the averaged BLEU scores on the first experience at the end of each subsequent one. SG-Rep exhibits more forgetting with respect to A-GEM but has a significantly better BLEU when considering the whole stream of experiences.
To investigate whether the order of the experiences has an effect on the different approaches compared, we ran additional experiments under 3 different permutations of the experiences.
We found SG-Rep to be quite resilient in this setting showing lower variations for the average BLEU score and $\Delta Lp$ than those exhibited by EWC and AGEM.
We report full data for these additional experiments in the Appendix \ref{sec:appendix-task-order-condensed}

\subsection{Sub-word Tokens Overlap}
In this particular setting, we observed that utilizing EWC regularization leads to subpar performance and fails to effectively mitigate catastrophic forgetting. We attribute this outcome to the inherent characteristics of this CL scenario, where the model must learn distinct target distributions that often exhibit minimal overlap, despite the utilization of a shared sub-word vocabulary.

\begin{table}[!h]
\centering
\resizebox{0.85\columnwidth}{!}{\begin{tabular}{c|cccc}
\hline
\multirow{2}{*}{\textbf{Exp.}} & \multicolumn{4}{c}{\textbf{Overlap \% of $k$ most frequent tokens}}                                        \\
                           & $k=10^2$                & $k=5\cdot10^2$                & $k=10^3$               & $k=5\cdot10^3$               \\
                           \hline
1-2                        & 20.20                & 26.45                & 26.22                & 26.74                \\
1-3                        & 57.57                & 57.51                & 55.65                & 55.47                \\
1-4                        & 16.16                & 22.22                & 23.52                & 26.02                \\
2-3                        & 17.17                & 27.65                & 28.22                & 27.74                \\
2-4                        & 50.50                & 55.71                & 56.65                & 58.75                \\
3-4                        & 47.47                & 58.31                & 58.95                & 61.43                \\
\hline
\end{tabular}}
\caption{Overlap percentage between the top-$k$ tokens of different experiences for several values of $k$.}
\label{tab:overlap}
\end{table}

To validate our hypothesis, we performed tokenization on the labels within each training set of the individual experiences. Subsequently, we computed the token frequencies and selected the top $k$ most frequent tokens for each experience. By measuring the intersection between these selected tokens, we quantified the overlap as a percentage. The results of this analysis, presented in Table \ref{tab:overlap}, highlight the degree of overlap for various values of $k$.
The findings demonstrate that while there is a higher overlap percentage between experiences that share the same target language, even for larger values of $k$, the overlap remains relatively low.

\begin{figure}[h]
    \centering
    \includegraphics[scale=0.45]{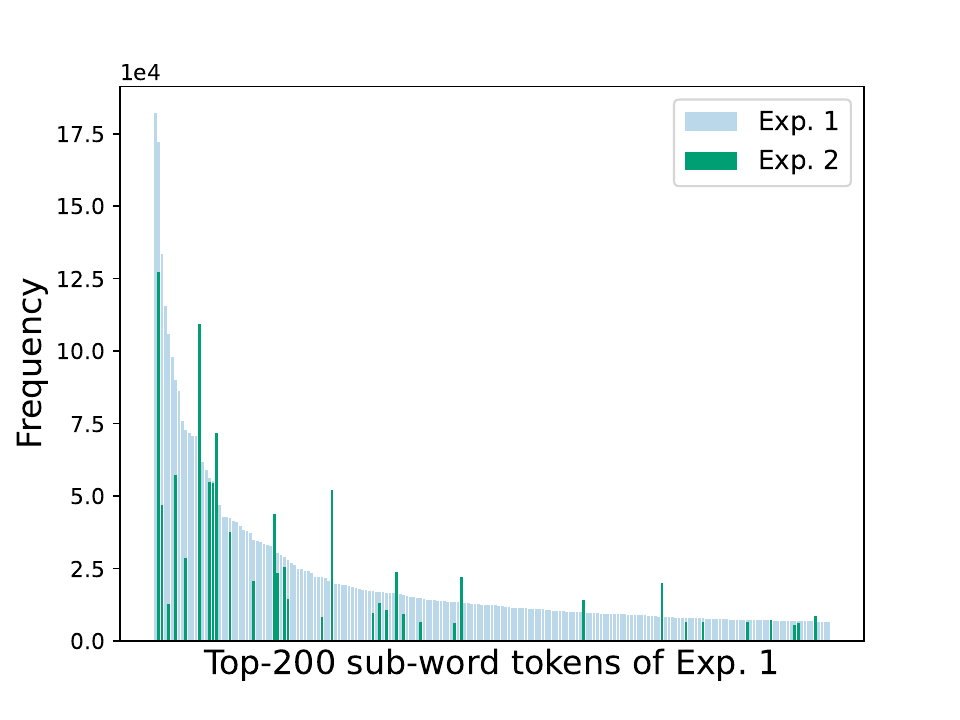}
    \caption{In light blue, frequencies of the top-200 sub-word tokens of exp. 1. The green bars represent the frequency of the same tokens that also appear in the 200 most frequent tokens of exp 2.}
    \label{fig:exp-overlap}
\end{figure}

For a visual insight into the token distribution, we provide Figure \ref{fig:exp-overlap}, which presents a plot comparing the top-200 sub-word tokens in Experience 1 with their occurrences in the top-200 tokens of Experience 2.
In the CILL setting, the token distribution affects the output layer in a manner akin to a class-incremental scenario,  albeit to a lesser extent.
Consequently, it is normal to expect a degradation of the performance: in such scenarios, EWC performs on par with fine-tuning \cite{ewc-class-incremental} and the effectiveness of regularization methods is generally acknowledged to be relatively low \cite{3-scenarios-cl}. 
Although A-GEM demonstrates a relatively smaller decline in performance compared to other methods, its performance remains suboptimal.
 Under a task order that simulates a Domain Incremental setting (e.g. with subsequent tasks sharing a language), EWC has a stronger performance. We report the scores for both EWC and AGEM in Appendix \ref{sec:appendix-task-order}. 

\begin{table*}[!h]
\centering
\resizebox{0.8\textwidth}{!}{\begin{tabular}{l|cc|cc|cc|cc|cc}
\hline
\textbf{Systems}  & \textbf{Ar$\rightarrow$En }     & \textbf{En$\rightarrow$Ar}      & \textbf{En$\rightarrow$Fr }      & \textbf{Fr$\rightarrow$En}       & \textbf{Ko$\rightarrow$En }      & \textbf{En$\rightarrow$Ko }     & \textbf{Nl$\rightarrow$It  }     & \textbf{It$\rightarrow$Nl }     & \textbf{Avg.} & $\mathbf{\Delta Lp^*} \downarrow $   \\
\hline
Incremental train & 0.01 & 0.06 & 0.48 & 0.56 & 0.03 & 0.16 & 21.29 & 21.59 &
5.52 & 17.60 \\
EWC & \textbf{29.79} & \textbf{12.37} & 1.15 & 14.68 & 2.64 & 0.38 &
0.07 & 0.11 & 7.65 & 15.47 \\
A-GEM$_{0.2}^{}$ & 13.8 & 3.91 & 0.83 & 0.63 & 0.17 & 0.25 & 20.73 &
\textbf{21.61} & 7.74 & 15.38 \\
SG-Rep$_{0.2}^{250}$ & 16.53 & 6.74 & \textbf{21.12} & \textbf{28.21}
& \textbf{9.65} & \textbf{3.41} & \textbf{21} & 21.47 & \textbf{16.02} &
\textbf{7.11} \\
\hdashline
Replay$_{0.1}$ & 26.53 & 10.19 & 33.7 & 35.19 & 14.66 & 4.83 & 20.85 &
21.38 & 20.92 & 2.21 \\
Joint training* & 30.98 & 12.7 & 37.69 & 39.5 & 17.29 & 5.74 & 20.33 &
23.12 & 20.55 & -- \\
\hline
\end{tabular}}
\caption{BLEU scores for experiments with non-European languages. The scores are computed at the end of the last experience (after training on Nl $\leftrightarrow$ It pair) on the corresponding IWSLT17 test sets. }
\label{tab:eastern}
\end{table*}

\subsection{Effects of The Different Hyper-parameters}
We investigated the impact of various hyper-parameters on the performance of our models.

In SG-Rep, we examined the effects of different memory sizes, specifically 5\%, 10\%, and 20\%. Additionally, we maintained a fixed memory size of 20\% and varied the number of self-generated samples, exploring values of $n = [100k, 180k, 250k]$.

\begin{figure}[]
    \centering
    \includegraphics[width=0.9\columnwidth]{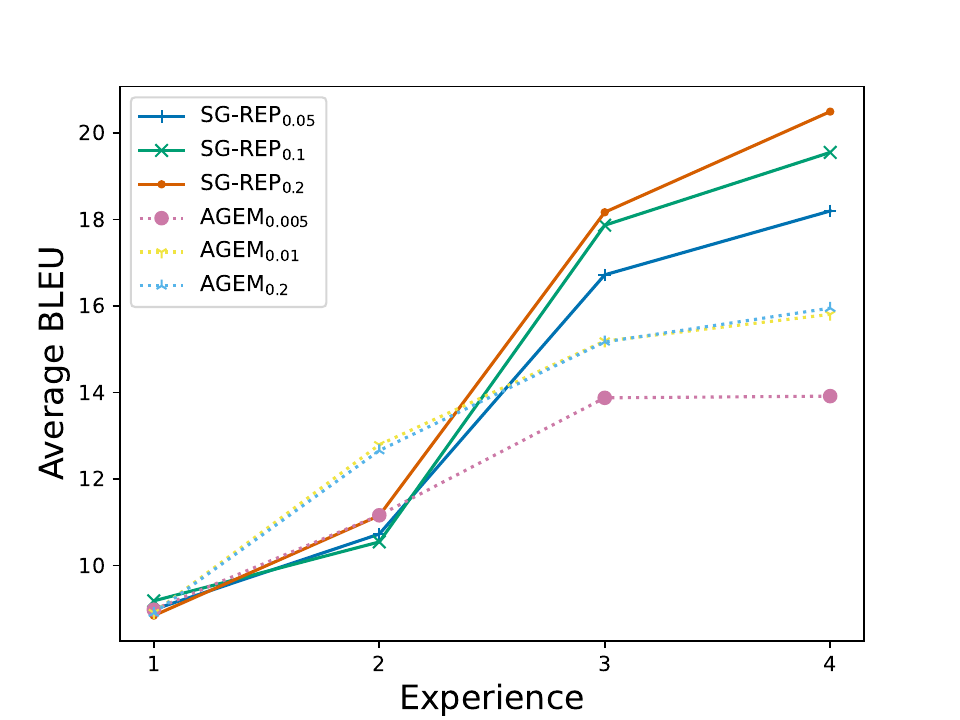}
    \caption{Effect of different memory sizes for A-GEM and SG-Rep.}
    \label{fig:memory size}
\end{figure}

For the A-GEM method, we utilized different memory sizes, namely 5\%, 10\%, and 20\% .
To investigate the impact of regularization strength, we varied the values of $\lambda$ for the EWC method, specifically using $\lambda = [0.25, 2, 200, 2000 ]$.
Figure \ref{fig:memory size} presents a comparison between A-GEM and our proposed approach across different memory sizes. Notably, increasing the memory size from 10\% to 20\% had a negligible effect on A-GEM, while it resulted in a relative improvement of nearly 1 BLEU for the self-generated replay approach.

\begin{table*}[!h]
\centering
\resizebox{0.8\textwidth}{!}{\begin{tabular}{l|cc|cc|cc|cc|cc}
\hline
\textbf{Systems} & \textbf{Ar$\rightarrow$En}      & \textbf{En$\rightarrow$Ar}      &\textbf{Es$\rightarrow$Ru }&\textbf{Ru$\rightarrow$Es} &\textbf{En$\rightarrow$Fr} &\textbf{Fr$\rightarrow$En} &\textbf{En$\rightarrow$Es} &\textbf{Es$\rightarrow$En} &\textbf{Avg.} &$\mathbf{\Delta Lp^*} \downarrow$ \\\hline
Incremental train &6.74 &	5.08&	6.42&	6.53	&7.85	&11.56	&49.93&	57.31&	18.93 &	27.01\\
EWC  &\textbf{53.84} &\textbf{36.31} &2.64 &5.05 &4.79 &9.26 &8.49 &9.02 &16.18 &29.76 \\
A-GEM$_{0.2}^{}$ &45.02 &22.81 &5.92 &5.03 &7.93 &11.92 &49.07 &57.32 &25.63 &20.31 \\
SG-Rep$_{0.2}^{250}$  &29.07 &13.53 &\textbf{19.39} &\textbf{30.81} &\textbf{16.42} &\textbf{15.78} &\textbf{49.95} &\textbf{57.45} &\textbf{29.05} &\textbf{16.89}\\
\hdashline
Replay$_{0.1}$  &39.84 &18.96 &26.55 &39.04 &22.4 &32.39 &47.04 &55.51 &35.22 &10.73 \\
Joint training*  &52.41 &34.9 &39.15 &44.82 &45.02 &50.77 &45.01 &55.46 &45.94 & --\\
\hline
\end{tabular}}
\caption{BLEU scores for experiments with UNPC. The scores are computed at the end of the last experience (after training on Es $\leftrightarrow$ En pair) on the corresponding UNPC test sets.}
\label{tab:unpc}
\end{table*}

\begin{figure}[h]
    \centering
    \includegraphics[width=0.9\columnwidth]{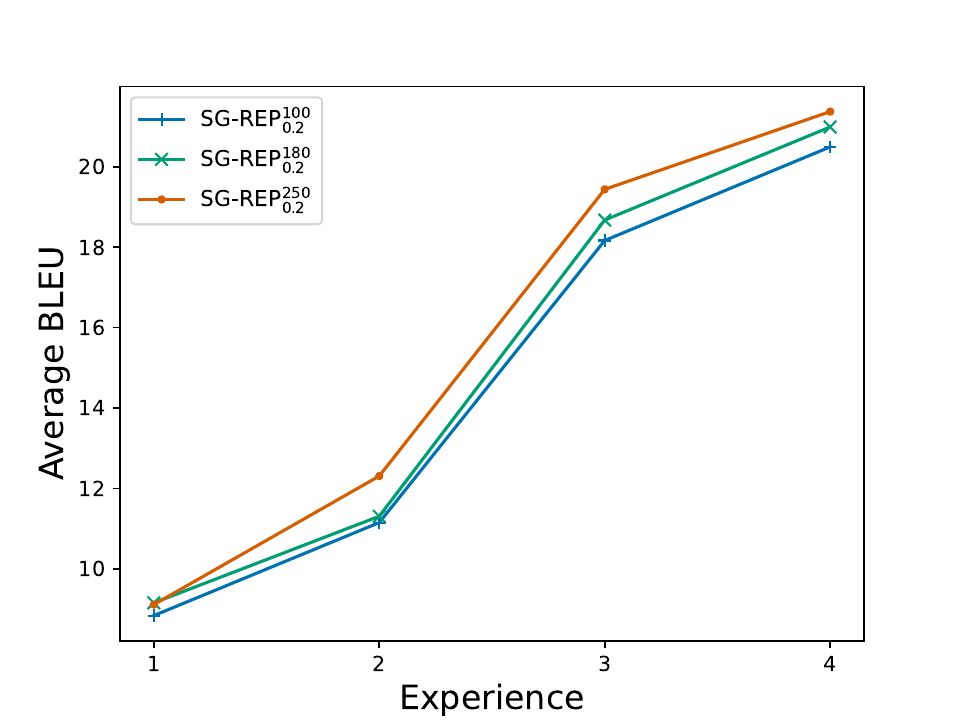}
    \caption{Average BLEU score on all language directions at the end of each training experience.}
    \label{fig:gen-samples-exp}
\end{figure}

By maintaining a fixed memory size and increasing the quantity of generated pseudo-samples, we obtain a larger initial population for reservoir sampling, leading to increased diversity within the memory. This diversity proved to be beneficial for the overall performance. Figure \ref{fig:gen-samples-exp} illustrates the average BLEU score at the conclusion of each experience, considering various values of self-generated samples while keeping a fixed replay buffer size of 20\%. 
We report a plot of the forgetting trend in Appendix \ref{appendix:gen-samples-forgetting}.
\subsection{IWSLT17 with Eastern Languages}
Table \ref{tab:eastern} summarizes 
 the results on the stream of experiences containing also non-European languages.
In this setting, both AGEM and EWC perform poorly, with the latter being able to retain most translation proficiency for Ar$\leftrightarrow$En but failing to learn in other directions. 
SG-Rep outperforms both AGEM and EWC by a larger margin with respect to the setting containing only European languages.

\begin{table*}[!h]
\centering
\resizebox{0.95\textwidth}{!}{\begin{tabular}{lccc|ccc|cc|c}
\hline
\multicolumn{1}{c}{\textbf{}}   & \multicolumn{1}{l}{} & \multicolumn{1}{l}{} & \multicolumn{2}{c}{\textbf{IWSLT17}}&    & \multicolumn{1}{l}{} & \multicolumn{1}{l}{} & \multicolumn{1}{l}{} & \multicolumn{1}{l}{}  \\[4pt]                                                                                                                                 
                                & \textbf{Fr-En} & \textbf{Nl-It} & \textbf{En-Ro} & \multicolumn{3}{c|}{\textbf{Fr-En}}         & \multicolumn{2}{c|}{\textbf{Nl-It}} & \textbf{En-Ro} \\
                                & Generated data & Generated data & Generated data & Buffer exp 1 & Buffer exp 2 & Buffer exp 3 & Buffer exp 2     & Buffer exp 3    & Buffer exp 3   \\
                                \hline
\% of leaked source          & 0.23 (623)     & 0.32 (887)     & 0.29 (760)     & 0.13 (487)   & 0.076 (276)  & 0.05 (211)   & 0.08 (302)       & 0.06 (227)      & 0.09 (325)     \\
\% of leaked target         & 0.26 (666)     & 0.30 (790)     & 0.26 (693)     & 0.19 (710)   & 0.14 (520)   & 0.11 (430)   & 0.20 (743)       & 0.174 (629)     & 0.18 (654)     \\
Avg. length of leaked source& 14.5           & 14.83          & 14.81          & 13.03        & 12.21        & 11.62        & 13.72            & 13.12           & 12.32          \\
Avg. length of leaked target & 13.64          & 13.31          & 11.79          & 11.78        & 1.51         & 11.14        & 12.66            & 11.98           & 9.9            \\
\hline
                                & \textbf{Ar-En} & \textbf{En-Fr} & \textbf{Ko-En} & \multicolumn{3}{c|}{\textbf{Ar-En}}         & \multicolumn{2}{c|}{\textbf{Fr-En}} & \textbf{Ko-En} \\
                                & Generated data & Generated data & Generated data & Buffer exp 1 & Buffer exp 2 & Buffer exp 3 & Buffer exp 2     & Buffer exp 3    & Buffer exp 3   \\
                                \hline
\% of leaked source          & 0.1 (264)      & 0.21 (560)     & 0.20 (557)     & 0.05 (208)   & 0.04 (153)   & 0.03 (123)   & 0.12 (461)       & 0.12 (449)      & 0.03 (136)     \\
\% of leaked target           & 0.24 (630)     & 0.005 (13)     & 0.30 (764)     & 0.15 (542)   & 0.11 (419)   & 0.09 (349)   & 0.13 (498)       & 0.21 (786)      & 0.088 (319)    \\
Avg. length of leaked source  & 7.70           & 12.73          & 6.88           & 6.92         & 6.451        & 6.15         & 12.67            & 13.13           & 6.74           \\
Avg. length of leaked target  & 13.04          & 4.08           & 14.34          & 6.5          & 6.39         & 5.77         & 8.74             & 11.14           & 4.83           \\[4pt]
\hline
\multicolumn{1}{c}{\textbf{}}   & \multicolumn{1}{l}{} & \multicolumn{1}{l}{} & \multicolumn{2}{c}{\textbf{UNPC}}&    & \multicolumn{1}{l}{} & \multicolumn{1}{l}{} & \multicolumn{1}{l}{} & \multicolumn{1}{l}{}                                                                                                                                             \\
                                & \textbf{Ar-En} & \textbf{Es-Ru} & \textbf{En-Fr} & \multicolumn{3}{c}{\textbf{Ar-En}}         & \multicolumn{2}{c}{\textbf{Es-Ru}} & \textbf{En-Fr} \\
                                & Generated data & Generated data & Generated data & Buffer exp 1 & Buffer exp 2 & Buffer exp 3 & Buffer exp 2     & Buffer exp 3    & Buffer exp 3   \\
                                \hline
\% of leaked source           & 5.63 (14517)   & 5.55 (14177)   & 6.54 (16607)   & 3.68         & 2.21         & 1.86         & 2.38             & 2.07            & 3.07           \\
\% of leaked target           & 5.74 (14635)   & 4.44 (11361)   & 7.12 (18011)   & 4.62         & 3.21         & 2.8          & 3.25             & 2.98            & 4.24           \\
Avg. length of leaked source  & 12.48          & 13.87          & 11.86          & 11.55        & 10.73        & 10.21        & 11.25            & 10.44           & 10.3          \\
Avg. length of leaked target  & 10.62          & 11.91          & 11.97          & 10.62        & 9.86         & 9.4          & 9.82             & 9.37            & 9.42   \\ \hline       
\end{tabular}}
\caption{Data leakage stats for IWSLT17 and UNPC datasets. Left: leakage in generated data before populating replay buffers. Right: data leaked into replay buffers. Stats are computed without mitigation.}
\label{tab:dataset-stats-leak}
\end{table*}

\subsection{UNPC}
We ran additional experiments on the United Nations Parallel Corpus in a high-resource context. Informed by the previous experiments on IWSLT and given the high computational costs we avoid to train low scoring approaches such as LAMOL. We keep EWC as an instance of regularization-based approaches. Table \ref{tab:unpc} summarizes the results.
EWC maintains the strongest performance on the Ar $\leftrightarrow$ En but impairs model learning in subsequent experiences. AGEM$_{0.2}$ is 
slightly better. SG-Rep has a larger gap with the classical Replay compared with the IWSLT setting, but it's still the strongest performer scoring more than 3 BLEU points higher than AGEM and having lower $\Delta Lp $ with the jointly trained model.

\subsection{Pseudo-samples Analysis}

We conducted an analysis of generated pseudo samples, covering duplicate counts and length statistics, to assess their characteristics and compare them with the original data. For diversity assessment, we utilized self-BLEU scores \cite{alihosseini-etal-2019-jointly}. Due to computational constraints, self-BLEU was calculated with a sampling approach, evaluating 5k sentences and reporting the average over 10 runs. Results are provided in Appendix \ref{appendix:true-vs-false} for both IWSLT17 data (Table \ref{tab:dataset-stats-iwslt}) and UNPC (Table \ref{tab:dataset-stats-unpc}). In general, generated data exhibit higher self-BLEU scores, indicating lower diversity compared to the original data, except for the Arabic-English language pair. Conversely, for UNPC, generated samples are more diverse than the original ones. Pseudo samples for both UNPC and IWSLT17 are shorter than the original data.

\subsection{Data leakage Analysis}
\label{sec:data-leak}
Table \ref{tab:dataset-stats-leak} summarizes the analysis quantifying training data leakage during model generation. For IWSLT17, the leaked data proportion is extremely low, below 0.5\%. It generally decreases for a single language pair when examining buffers and across experiences. English leakage consists of very short, common sentences with simple grammatical construction, like "Thank you" or "I am."
Comparatively, UNPC has a larger data leakage, but the percentage remains low. Also for UNPC the leaked sentences are short and frequent names (e.g., country names) and section titles (e.g., "Introduction," "B. Text"). 

\section{Conclusions}
We proposed a simple yet effective continual learning method for NMT that uses a replay memory to mitigate catastrophic forgetting. Differently from other data-based approaches, we do not memorize training samples explicitly and instead use the model itself as a generator of parallel sentences.
The experimental results prove that our method can achieve significant improvement over several strong continual learning baselines.

\paragraph{Limitations.}
Our method has a computational overhead to standard replay due to pseudo-sample generation. However, the overall training time for a stream of experiences is comparable to other baselines.
To assess the performance of SG-Rep in challenging scenarios,  we choose a model architecture with a relatively low number of parameters compared to state-of-the-art M-NMT systems.


\paragraph{Ethics Statement.}
Our work pertains to the continual training of NMT systems to adapt them with low forgetting. In this work, we use only publicly available data.

\paragraph{Acknowledgement}
Work supported by the EU NextGenerationEU programme under the funding scheme PNRR-PE-AI (PE00000013) FAIR - Future Artificial Intelligence Research (spoke 1). 

\bibliography{anthology,custom}
\bibliographystyle{acl_natbib}

\newpage

\onecolumn
\appendix

\section{Self-Generated Sentences}
\label{appendix:self-gen-samples}

\begin{table}[h]
\centering
\begin{tabular}{|l|l|}
\hline
\multicolumn{2}{|c|}{\textbf{Self-Generated Sentences}}                      \\
\hline
\textbf{English}                                        & \textbf{Dutch} \\
\hline
\textit{It has been something that most of them do.}    & \textit{Ik zou het kunnen zijn..}                      \\
\textit{It kind of leads in the back of the way}        & \textit{Ik hoorde het, oké, maar ik was er niet bij.}   \\
\textit{It's counterintuitive and heart failure. }      & \textit{Waarom?? Het was niet zo?}                      \\
\textit{They would turn to their concretes.}            & \textit{Dus ik ga het hebben over een bepaald soort ding.}\\
\textit{At least the very same thing.}                  &                                                 \\
\hline
\textbf{French}   & \textbf{Italian} \\
\hline
\textit{puisqu'il s'agit de construire des courantes.}           & \textit{È fantastico.} \\
\textit{Et voici ce qu'il faut.}                                 & \textit{Così ho fatto una cosa come il gruppo di bambini} \\
\textit{Leurs biens étaient engendrés par le passé.}             & \textit{Eppure, si tratta di piccole specie.}\\
\textit{C'est simplement la homme qui fait le charge.}           & \textit{Quelli che lo provoquono.} \\
\hline
\textbf{Romanian} & \textbf{Filtered Sentences in English} \\
\hline
\textit{Arena Pahăriări!'} & \textit{It turns out, \underline{ates} are "w \underline{gravk}."}\\ 
\textit{Iată.}             & \textit{It's called a c \underline{l'homme-afour}.}\\       
\textit{Cu Cuvântul!'}     & \textit{\underline{fourn fourn économied} the question --}\\
\textit{Wețea with that}   & \textit{It's sort \underline{ofky knifery} to us.}\\ 

\hline
\end{tabular}
\caption{Generated samples for several languages and filtered-out English sentences (bottom right) in the self-generation process. Underlined words indicate errors detected by PyEnchant.}
\end{table}

\section{ IWSLT17 Score Under Different Task Order}
\label{sec:appendix-task-order-condensed}

\begin{table*}[h]
\centering
\resizebox{\textwidth}{!}{\begin{tabular}{l|cc|cc|cc|cc|cc}
\hline
\multirow{2}*{Systems} &\multicolumn{8}{c|}{Permutation 1} & \multirow{2}*{\textbf{Avg.}} & \multirow{2}*{$\mathbf{\Delta Lp^*} \downarrow $} \\
& \textbf{It$\rightarrow$Ro}      & \textbf{Ro$\rightarrow$It}      & \textbf{Fr$\rightarrow$En}       & \textbf{En$\rightarrow$Fr}      & \textbf{Nl$\rightarrow$It}       & \textbf{It$\rightarrow$N}l      &  \textbf{En$\rightarrow$Ro}       & \textbf{Ro$\rightarrow$En}   &     &   \\
\hline
\hline

EWC & \textbf{20.32} & \textbf{22.29} & 13.1 & 4.74 & 3.92 & 0.73 & 0.15
& 0.92 & 8.27 & 20.97 \\[3pt]
A-GEM$_{0.2}^{}$ & 15.99 & 18.02 & 10.74 & 24.65 & 1.65 & 1.12 &
\textbf{40.88} & \textbf{39.96} & 19.12 & 10.11 \\[3pt]
SG-Rep$_{0.2}^{250}$  & 15.14 & 17.62 & \textbf{18.96} &
\textbf{28.99} & \textbf{16.98} & \textbf{15.7} & 40.77 & 39.62 &
\textbf{24.22} & \textbf{5.02} \\[3pt]
\hline

& \multicolumn{8}{c|}{Permutation 2} & \multicolumn{2}{c}{}\\

  &  \textbf{En$\rightarrow$Ro}       & \textbf{Ro$\rightarrow$En}   & \textbf{It$\rightarrow$Ro}      & \textbf{Ro$\rightarrow$It}      & \textbf{Fr$\rightarrow$En}       & \textbf{En$\rightarrow$Fr}      & \textbf{Nl$\rightarrow$It}       & \textbf{It$\rightarrow$Nl}      &      & \\
\hline
\hline
EWC & \textbf{20.81} & \textbf{22.36} & 4.14 & 3.23 & 1.28 & 1.01 & 1.53
& 0.91 & 6.90 & 22.33 \\[3pt]
A-GEM$_{0.2}^{}$ & 18.7 & 19.0 & 4.55 & 1.13 & 3.47 & 0.99 &
\textbf{28.41} & \textbf{35.6} & 13.98 & 15.26 \\[3pt]
SG-Rep$_{0.2}^{250}$  & 18.4 & 18.23 & \textbf{34.08} & \textbf{28.98}
& \textbf{16.31} & \textbf{15.87} & 27.51 & 35.11 & \textbf{24.31} &
\textbf{4.93} \\[3pt]
\hline
\hline
& \multicolumn{8}{c|}{Permutation 3} & \multicolumn{2}{c}{}\\
&  \textbf{En$\rightarrow$Ro}       & \textbf{Ro$\rightarrow$En}   & \textbf{It$\rightarrow$Ro}      & \textbf{Ro$\rightarrow$It}      & \textbf{Fr$\rightarrow$En}       & \textbf{En$\rightarrow$Fr}      & \textbf{Nl$\rightarrow$It}       & \textbf{It$\rightarrow$Nl}      &      & \\
\hline
\hline
EWC & 7.67 & 10.48 & 2.05 & 0.48 & 2.91 & 0.87 & 0.04 & 0.35 & 3.10&
26.13 \\[3pt]
A-GEM$_{0.2}^{}$ & \textbf{20.23} & \textbf{26.37} & 3.06 & 14.31 &
7.76 & 1.09 & 21.77 & 21.99 & 14.57 & 14.67 \\[3pt]
SG-Rep$_{0.2}^{250}$ & 19.96 & 25.63 & \textbf{14.73} & \textbf{19} &
\textbf{30.38} & \textbf{24.5} & \textbf{21.69} & \textbf{22.13} &
\textbf{22.25} & \textbf{6.99} \\[3pt]
\hline
Joint training* & 28.13 & 35.11 & 22.10 & 23.58 & 40.90 & 39.77 & 22.22
& 22.10 & 29.24 & -\/- \\
\hline
\end{tabular}}
\caption{Performance scores on IWSLT17 test set computed at the end of the fourth experience under different task permutations.}
\label{tab:iwslt17-perm-euro}
\end{table*}

\newpage

\section{AGEM and EWC Scores Different Task Order}
\label{sec:appendix-task-order}

\begin{table*}[h]
\centering
\begin{tabular}{ccc|cc|cc|c}
\hline
\textbf{Exp.} &    \textbf{Fr$\rightarrow$En }&    \textbf{En$\rightarrow$Fr} &    \textbf{En$\rightarrow$Ro}       & \textbf{Ro$\rightarrow$En} &    \textbf{It$\rightarrow$Ro}       & \textbf{Ro$\rightarrow$It} &    \textbf{Avg.} \\
\hline
1 &  39.49 &  38.07 &   0.95 &   0.85 &   0.45 &   0.18 &  13.33 \\
2 &  33.50 &  27.70 &  27.98 &  34.36 &   1.01 &   0.74 &  20.88 \\
3 &  29.22 &  26.16 &  22.06 &   8.33 &  21.14 &  23.40 &  21.71 \\
\hline
\end{tabular}
\caption{A-GEM$_{0.1}$ performance under a different experience order. Training data of the various experiences is from IWSLT17 dataset.}
\end{table*}

\begin{table*}[h]
\centering
\begin{tabular}{ccc|cc|cc|c}
\hline
\textbf{Exp.} &   \textbf{ Fr$\rightarrow$En }&    \textbf{En$\rightarrow$Fr} &    \textbf{En$\rightarrow$Ro}       & \textbf{Ro$\rightarrow$En} &    \textbf{It$\rightarrow$Ro}       & \textbf{Ro$\rightarrow$It} &    \textbf{Avg.} \\
\hline
1 &  39.63 &  38.32 &   1.04 &   0.54 &   0.18 &   0.13 &  13.30 \\
2 &  28.43 &   1.15 &  27.14 &  34.01 &   0.89 &   0.82 &  15.40 \\
3 &   1.22 &   1.07 &  22.04 &   0.92 &  20.19 &  21.48 &   11.15 \\
\hline
\end{tabular}
\caption{EWC performance under a different experience order. Training data of the various experiences is from the IWSLT17 dataset. $\lambda=2k$}
\end{table*}

\newpage

\section{Sizes of the Replay Buffer}
\label{appendix:mem-size}

\begin{table*}[h]
\centering
\begin{tabular}{cr}
\textbf{Memory size }& \textbf{Num. samples} \\
\hline
       5\%           &    90433    \\
       10\%         &    180866   \\
       20\%        &    361732         \\
       \hline
       100\%        &   1,808,658\\
\hline
\end{tabular}
\caption{Total sizes of the replay buffer. A size of 100\% is the sum of all the training samples across the four experiences created out of the IWSLT17 dataset (European-only languages). }
\end{table*}

\section{Forgetting curve for different sizes of generated samples}
\label{appendix:gen-samples-forgetting}

\begin{figure}[h]
    \centering
    \includegraphics[width=0.5\columnwidth]{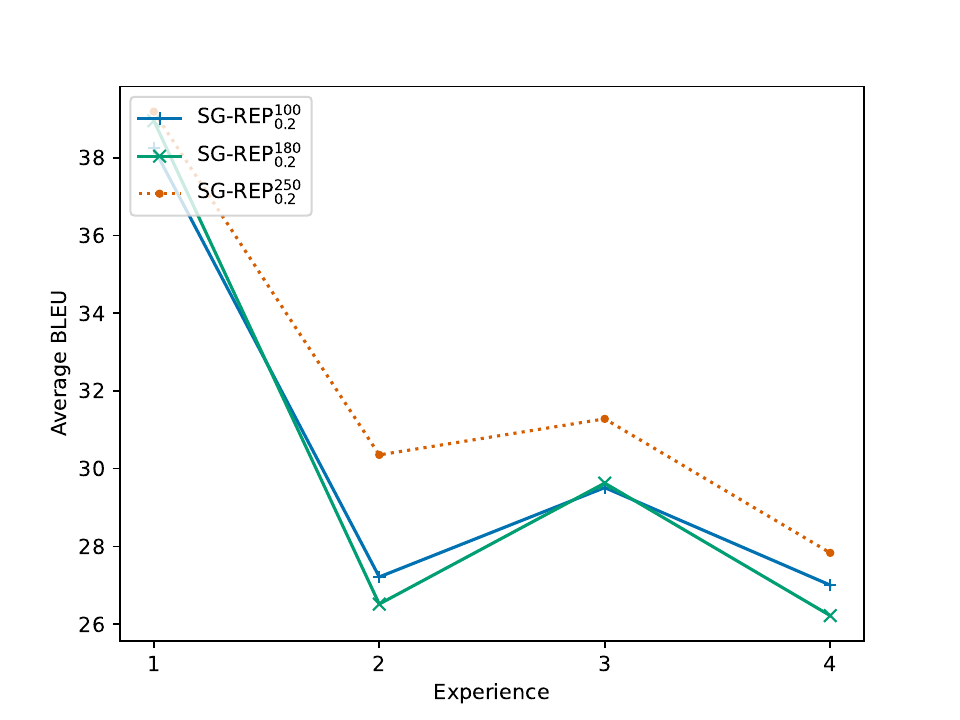}
    \caption{Forgetting curve of SG-Rep for different numbers of self-generated samples. Average BLEU score on the first task evaluated at the end of the training process of each experience.}
\end{figure}

\section{Training time of the different methodologies}

\begin{table}[h]
\centering
\begin{tabular}{lcc}
\hline
            & \multicolumn{2}{c}{\textbf{Time (h) for Dataset}}          \\
\textbf{Method}      & \textbf{UNPC }   & \textbf{IWSLT17 non-Europ. } \\
\hline
Incremental & 62.45    & 6.21    \\
EWC         & 128.13  & 29.08   \\
AGEM        & 67.81   & 14.43   \\
SG-Rep      & 64.03   & 11.74   \\
Replay      & 57.66   & 14.84   \\
Joint       & 62.66   & 11.94  \\
\hline
\end{tabular}
\caption{Total training time measured in hours for the different approaches. SG-Rep is faster than AGEM and EWC and very close to pure replay. }
\end{table}

\newpage

\section{COMET Scores }
\label{appendix:comet-scores}

\begin{table*}[!h]
\centering
\resizebox{\textwidth}{!}{\begin{tabular}{l|cc|cc|cc|cc|cc}
\hline

\textbf{Systems } & \textbf{Fr$\rightarrow$En }     & \textbf{En$\rightarrow$Fr}      & \textbf{Nl$\rightarrow$It }      & \textbf{It$\rightarrow$Nl}       & \textbf{En$\rightarrow$Ro}       & \textbf{Ro$\rightarrow$En }     & \textbf{It$\rightarrow$Ro }      & \textbf{Ro$\rightarrow$It}      & \textbf{Avg.}  & $\mathbf{\Delta Lp^*} \downarrow $   \\
\hline
\hline
Incremental train & 0.33  & 0.51  & 0.33  & 0.67  & 0.69  & 0.7   & 0.82  & 0.81  & 0.61 & 0.20               \\
EWC & \textbf{0.83}  & \textbf{0.79}  & 0.37  & 0.39  & 0.56  & 0.53  & 0.23  & 0.3   & 0.5  & 0.31               \\
A-GEM$_{0.2}$   & 0.75  & 0.69  & 0.32  & 0.67  & 0.76  & 0.69  & 0.82  & 0.81  & 0.69 & 0.12               \\
SG-Rep$_{0.2}^{250}$ & 0.75  & 0.68  & \textbf{0.71}  & \textbf{0.68}  & \textbf{0.79 } & \textbf{0.77}  & \textbf{0.82}  & \textbf{0.81 } & \textbf{0.75} & \textbf{0.06}               \\
\hdashline
Replay$_{0.1}$  & 0.81  & 0.76  & 0.74  & 0.74  & 0.82  & 0.82  & 0.82  & 0.81  & 0.79 & 0.02               \\
Joint training* & 0.84  & 0.8   & 0.77  & 0.78  & 0.83  & 0.84  & 0.82  & 0.81  & 0.81 & -- \\
\hline
\end{tabular}}
\caption{ COMET scores for experiments on IWSLT17 with non-European languages. Scores are computed at the end of the last experience. The Avg column is the average COMET score across all translation directions. Joint training and Replay are shown as upper bounds at the bottom.}
\label{tab:main_results-comet}
\end{table*}

\begin{table*}[!h]
\centering
\resizebox{\textwidth}{!}{\begin{tabular}{l|cc|cc|cc|cc|cc}
\hline
\textbf{Systems}  & \textbf{Ar$\rightarrow$En }     & \textbf{En$\rightarrow$Ar}      & \textbf{En$\rightarrow$Fr }      & \textbf{Fr$\rightarrow$En}       & \textbf{Ko$\rightarrow$En }      & \textbf{En$\rightarrow$Ko }     & \textbf{Nl$\rightarrow$It  }     & \textbf{It$\rightarrow$Nl }     & \textbf{Avg.} & $\mathbf{\Delta Lp^*} \downarrow $   \\
\hline
Incremental          & 0.23 & 0.28 & 0.31 & 0.29 & 0.29 & 0.33 & 0.77 & 0.78 & 0.41 & 0.32    \\
EWC                  & \textbf{0.77} &\textbf{ 0.78} & 0.27 & 0.57 & 0.46 & 0.24 & 0.25 & 0.26 & 0.45 & 0.28    \\
A-GEM$_{0.2}^{}$     & 0.69 & 0.69 & 0.52 & 0.33 & 0.3  & 0.46 & 0.77 & 0.78 & 0.57 & 0.16   \\
SG-Rep$_{0.2}^{250}$ & 0.66 & 0.64 & \textbf{0.61} & \textbf{0.74} & \textbf{0.67} & \textbf{0.67} & \textbf{0.77} & \textbf{0.78} & \textbf{0.69} & \textbf{0.04}   \\
\hdashline
Replay$_{0.1}$        & 0.74 & 0.75 & 0.76 & 0.8  & 0.73 & 0.77 & 0.77 & 0.77 & 0.76 & -0.02 \\
Joint training*       & 0.73 & 0.74 & 0.75 & 0.8  & 0.72 & 0.74 & 0.69 & 0.71 & 0.73 & --     \\

\hline
\end{tabular}}
\caption{COMET scores for experiments on IWSLT17 with non-European languages. The scores are computed at the end of the last experience (after training on Nl $\leftrightarrow$ It pair) on the corresponding test sets.
Joint training and Replay are shown as upper bounds at the bottom.}
\label{tab:eastern-comet}
\end{table*}

\begin{table*}[!h]

\centering
\resizebox{\textwidth}{!}{\begin{tabular}{l|cc|cc|cc|cc|cc}
\hline
\textbf{Systems} & \textbf{Ar$\rightarrow$En}      & \textbf{En$\rightarrow$Ar}      &\textbf{Es$\rightarrow$Ru }&\textbf{Ru$\rightarrow$Es} &\textbf{En$\rightarrow$Fr} &\textbf{Fr$\rightarrow$En} &\textbf{En$\rightarrow$Es} &\textbf{Es$\rightarrow$En} &\textbf{Avg.} &$\mathbf{\Delta Lp^*} \downarrow$ \\\hline
Incremental train & 0.61 & 0.53 & 0.5  & 0.73 & 0.74 & 0.77 & 0.87 & 0.89 & 0.7  & 0.12 \\
EWC         & \textbf{0.87} & \textbf{0.84} & 0.45 & 0.3  & 0.6  & 0.76 & 0.63 & 0.73 & 0.65 & 0.18 \\
AGEM$_{0.2}$        & 0.77 & 0.63 & 0.51 & 0.73 & 0.73 & 0.77 & 0.86 & 0.89 & 0.74 & 0.09  \\
SG-Rep$_{0.2}^{250}$       & 0.76 & 0.56 & \textbf{0.56} & \textbf{0.77} & \textbf{0.74 }& \textbf{0.8}  & \textbf{0.88} & \textbf{0.9 } & \textbf{0.74 }& \textbf{0.08}  \\
\hdashline
Replay$_{0.1}$       & 0.77 & 0.61 & 0.66 & 0.79 & 0.73 & 0.81 & 0.87 & 0.89 & 0.77 & 0.06  \\
Joint training*     & 0.83 & 0.81 & 0.83 & 0.82 & 0.81 & 0.86 & 0.84 & 0.87 & 0.83 & --    \\
\hline
\end{tabular}}
\caption{COMET scores for experiments with UNPC. The scores are computed at the end of the last experience (after training on Es $\leftrightarrow$ En pair) on the corresponding UNPC test sets.Joint training and Replay are shown as upper bounds at the bottom.}
\label{tab:unpc-comet}
\end{table*}

\newpage

\section{Statistics of Original and Pseudo-samples}
\label{appendix:true-vs-false}

We conducted an analysis of generated pseudo samples, covering duplicate counts and length statistics, to assess their characteristics and compare them with the original data. 

For diversity assessment, we utilized self-BLEU scores \cite{alihosseini-etal-2019-jointly}. Due to computational constraints, self-BLEU was calculated with a sampling approach, evaluating 5k sentences and reporting the average over 10 runs. Results are provided for both IWSLT17 data in Table\ref{tab:dataset-stats-iwslt} and UNPC in Table \ref{tab:dataset-stats-unpc}. 

In general, generated data exhibit higher self-BLEU scores, indicating lower diversity compared to the original data, except for the Arabic-English language pair. Conversely, for UNPC, generated samples are more diverse than the original ones. Pseudo samples for both UNPC and IWSLT17 are shorter than the original data.

\begin{table*}[h]
\resizebox{\textwidth}{!}{\begin{tabular}{lcc|cc|cc|cc|cc}
\hline
                          & \multicolumn{2}{c|}{\textbf{Fr-En}} & \multicolumn{2}{c|}{\textbf{Nl-It}} & \multicolumn{2}{c|}{\textbf{En-Ro}} & \multicolumn{2}{c|}{\textbf{Ar-En}} & \multicolumn{2}{c}{\textbf{Ko-En}} \\
                          & Original data    & Generated       & Original         & Generated       & Original         & Generated       & Original data    & Generated       & Original         & Generated       \\
                          \hline
Avg. length source & 105.62 $\pm$ 74.88  & 24.74 $\pm$ 12.15  & 85.94 $\pm$ 58.63    & 26.82 $\pm$ 13.39   & 94.09 $\pm$ 66.62    & 23.04 $\pm$ 10.88   & 77.89 $\pm$ 55.93    & 25.10 $\pm$ 15.92   & 49.60 $\pm$ 34.38  & 12.587 $\pm$ 6.37  \\
Min. source length & 1                & 1               & 1                & 1               & 1                & 1               & 1                & 1               & 1                & 1               \\
Max. source length & 557              & 265             & 547              & 184             & 523              & 264             & 490              & 262             & 357              & 294             \\
Duplicated source sents & 2887             & 0               & 3666             & 0               & 2474             & 0               & 2032             & 0               & 2286             & 0               \\
Avg. Self-BLEU  source & 23.97            & 21.02           & 18.68            & 21.08           & 26.48            & 47.67           & 10.9             & 15.93           & 11.44            & 14.04           \\
\hdashline
Avg. length target & 94.95 $\pm$ 67.34  & 24.15 $\pm$ 12.09 & 89.45 $\pm$ 61.28  & 28.64 $\pm$ 14.78 & 92.55 $\pm$ 66.06  & 22.74 $\pm$ 11.43  & 94.96 $\pm$ 67.8     & 14.93 $\pm$ 12.21   & 95.02 $\pm$ 68.12   & 28.95 $\pm$ 21.12  \\
Min. target length & 1                & 1               & 1                & 1               & 1                & 1               & 1                & 1               & 3                & 1               \\
Max. source length & 523              & 256             & 562              & 174             & 556              & 258             & 514              & 886             & 531              & 895             \\
Avg. Self-BLEU  target & 26.52            & 28.02           & 18.33            & 27.56           & 18.14            & 36.22           & 26.5             & 20.45           & 26.42            & 40.77           \\
Duplicated target sents  & 2827             & 20502           & 2582             & 20681           & 2431             & 32326           & 2836             & 14418           & 2816             & 40052        \\
\hline
\end{tabular}}
\caption{Comparison of dataset statistics between language pairs in IWSLT17 and generated pseudo samples. Avg. Self-BLEU represents the average over 10 runs with a sample size of 5k sentences.}
\label{tab:dataset-stats-iwslt}
\end{table*}

\begin{table}[h]
\resizebox{\textwidth}{!}{\begin{tabular}{lcccccc}
\hline
                        & \multicolumn{2}{c}{\textbf{Ar-En}} & \multicolumn{2}{c}{\textbf{Es-Ru}} & \multicolumn{2}{c}{\textbf{En-Fr}}  \\
                        & Original data     & Generated      & Original          & Generated      & Original          & Generated       \\
                        \hline
Avg. length source      & 132.42 ± 112.08   & 19.66 ± 15.75  & 171.20 ± 147.66   & 23.39 ± 22.91  & 140.747 ± 126.661 & 24.253 ± 23.53  \\
Min. source length      & 1                 & 1              & 1                 & 1              & 1                 & 1               \\
Max. source length      & 15523             & 641            & 14164             & 684            & 15024             & 601             \\
Duplicated source sents & 3828415           & 0              & 5086430           & 0              & 8696252           & 0               \\
Avg. Self-BLEU  source  & 16.37             & 14.59          & 34.98             & 17.7           & 26.86             & 14.02           \\
\hdashline
Avg. length target      & 157.28 ± 132.64   & 20.19 ± 18.70  & 166.97 ± 144.29   & 21.98 ± 22.54  & 166.41 ± 151.952  & 27.492 ± 28.745 \\
Min. target length      & 1                 & 1              & 1                 & 1              & 1                 & 1               \\
Max. source length      & 15024             & 655            & 13684             & 800            & 22285             & 669             \\
Avg. Self-BLEU  target  & 29.46             & 16.73          & 21.83             & 15.1           & 40.19             & 20.97           \\
Duplicated target sents & 4517013           & 35608          & 5053606           & 50771          & 8276589           & 17988          \\
\hline
\end{tabular}}
\caption{Comparison of dataset statistics between language pairs in UNPC and generated pseudo samples. Avg. Self-BLEU represents the average over 10 runs with a sample size of 5k sentences.}
\label{tab:dataset-stats-unpc} 
\end{table}

\newpage

\end{document}